\documentclass[10pt, a4paper]{article}
\usepackage{booktabs}
\usepackage{tabularx}
\usepackage{todonotes}
\usepackage[final]{lrec2026} 
\usepackage[title]{appendix}
\newcolumntype{C}{>{\centering\arraybackslash}X}

\title{BIT.UA-AAUBS at ArchEHR-QA 2026: Evaluating Open-Source and Proprietary LLMs via Prompting in Low-Resource QA}

\name{Richard A. A. Jonker$^{\ast\dagger}$, Alexander Christiansen$^{\ast}$, Alexandros Maniatis$^{\ast}$, \\ \large \textbf{Rúben Garrido$^{\dagger}$, Rogério Braunschweiger de Freitas Lima$^{\ast}$,} \\ \large \textbf{Roman Jurowetzki$^{\ast}$, and Sérgio Matos$^{\dagger}$}}

\address{$^{\ast}$Aalborg University Business School \\ Fibigerstræde 2, Aalborg East 9220, Denmark \\ 
         \{raaj, ach22, amania24, rlima23, roman\}@business.aau.dk \\ \\
         $^{\dagger}$IEETA, DETI, LASI, University of Aveiro \\ Campus Universitário de Santiago, Aveiro 3810-193, Portugal \\ 
         \{richard.jonker, rubengarrido, aleixomatos\}@ua.pt }

\abstract{
This paper presents the joint participation of the BIT.UA and AAUBS groups in the ArchEHR-QA 2026 shared task, which focuses on clinical question answering and evidence grounding in a low-resource setting. Due to the absence of training data and the strict data privacy constraints inherent to the healthcare domain (e.g. GDPR), we investigate the capabilities of Large Language Models (LLMs) without weight updates. We evaluate several state-of-the-art proprietary models and locally deployable open-source alternatives using various prompt engineering strategies, including task decomposition, Chain-of-Thought, and in-context learning. Furthermore, we explore majority voting and LLM-as-a-judge ensembling techniques to maximize predictive robustness. Our results demonstrate that while proprietary models exhibit strong resilience to prompt variations, domain-adapted open-source models (such as MedGemma 3 27B) achieve highly competitive performance when paired with the right prompt. Overall, our prompt-based approach proved highly effective, securing 1st place in Subtask 4 (evidence citation alignment) and 3rd place in Subtask 3 (patient-friendly answer generation). All code, results, and prompts are available on our GitHub repository: \url{https://github.com/bioinformatics-ua/ArchEHR-QA-2026}. 
\\ \newline  }

\begin{document}

\maketitleabstract

\section{Introduction}

Patients increasingly seek to understand their health conditions and clinical course by reviewing their electronic health records (EHRs). However, clinical notes are notoriously complex, lengthy, and filled with medical jargon, making it difficult for patients to extract clear, accurate answers to their questions. The ArchEHR-QA 2026 \citep{soni-etal-2026-archehr-qa, soni-demner-fushman-2026-dataset} Shared Task  addresses this by challenging systems to perform grounded question answering (QA) directly from patient-specific EHRs. Unlike general health QA, this task emphasizes that clinical grounding ensures generated answers are explicitly linked to supporting evidence within the clinical notes.

The ArchEHR-QA 2026 challenge is divided into four sequential Subtasks: interpreting a verbose patient question into a concise clinical query (Subtask 1), identifying relevant evidence sentences from the EHR (Subtask 2), generating a patient-friendly answer (Subtask 3), and aligning the generated answer with the underlying clinical evidence (Subtask 4). Developing supervised models for these highly specialized steps is difficult due to the lack of annotated data.

To overcome the lack of extensive training data, our team participated in all four Subtasks by exploring a purely generative approach. We evaluated a wide variety of Large Language Models (LLMs), encompassing both open-source and proprietary architectures, relying entirely on prompting strategies. Our goal was to determine the extent to which off-the-shelf LLMs can bridge the semantic gap between patient inquiries and EHR data across the entire QA pipeline. Specifically, this study addresses two core research questions:
\begin{itemize}
    \item \textbf{RQ1:} To what extent can open-source LLMs bridge the performance gap to state-of-the-art proprietary models on complex clinical reasoning tasks?
    \item \textbf{RQ2:} To what extent does prompt engineering impact model performance and how does this vary between models.
\end{itemize}

Our comprehensive evaluation revealed substantial variability across the four Subtasks, highlighting the uneven capabilities of current LLMs in clinical contexts. While our systems achieved top results in the generation and alignment tasks, securing 1st place in Subtask 4 and 3rd place in Subtask 3, performance dropped considerably in the pure extraction and query formulation tasks. The remainder of this paper is structured as follows: Section 2 outlines the background and related work. Section 3 details our methodology, including our experimental setup and the diverse prompting techniques deployed. Section 4 presents our validation and official competition results, with Section 5 presenting some error analysis. Section 6 provides a  discussion with Section 7 concluding our findings.

\section{Background}

\textbf{Large Language Models in Clinical NLP.}
Until recently, state-of-the-art clinical natural language processing relied heavily on domain-specific, encoder-only architectures, such as ClinicalBERT \cite{clinicalbert}, which required extensive supervised fine-tuning. Recently, the paradigm has shifted toward generative LLMs. Models such as GPT-4, Gemini, and Claude have demonstrated remarkable zero-shot and few-shot capabilities across complex medical tasks, including clinical summarization, diagnostic reasoning, and patient-friendly translation \cite{thirunavukarasu2023large, singhal2023large}. However, applying these models to EHRs introduces significant challenges regarding hallucination \cite{huang2025survey} and factual consistency, necessitating strict grounding mechanisms to ensure patient safety.

\textbf{Prompt Engineering and In-Context Learning.}
While supervised fine-tuning remains the conventional standard for clinical NLP tasks, it is highly susceptible to overfitting in extreme low-resource settings and often impractical without extensive synthetic data generation \citep{meng2023tuning}. Consequently, advanced prompt engineering has emerged as the primary method for steering LLM behavior without weight updates \citep{sahoo2024systematic, chen2025unleashing}. Foundational techniques like few-shot In-Context Learning (ICL) \cite{brown2020language} and strict lexical constraints \cite{schall-de-melo-2025-hidden} help anchor model outputs to specific source texts. For complex clinical reasoning tasks, decomposing instructions via Task Decomposition \cite{zhou2023leasttomost} and Chain-of-Thought (CoT) prompting \cite{wei2022cot} have proven highly effective at imporving general performance. Furthermore, ensemble strategies, such as ``LLM-as-a-judge'' framework \cite{zheng2023judging}, are increasingly utilized to smooth out the variance inherent to generative models, improving both the robustness and precision of structured outputs.

\textbf{Privacy Constraints and Open-Source Alternatives.}
A primary barrier to deploying proprietary LLMs in real-world clinical environments is patient data privacy. Regulations such as the General Data Protection Regulation (GDPR) \cite{gdpr2016} and the Health Insurance Portability and Accountability Act (HIPAA) \cite{hipaa1996} often strictly prohibit the transmission of EHR data to external third-party APIs \citep{thirunavukarasu2023large}. Consequently, there is a growing imperative to develop and evaluate locally deployable, open-weights models. Recent releases such as MedGemma \cite{sellergren2025medgemma}, attempt to bridge the performance gap between open-source and closed-source systems specifically in the medical domain. Evaluating the true viability of these open-weights models against state-of-the-art proprietary baselines remains a critical area of ongoing research.

\section{Methodology}

Given the extreme low-resource constraints of this competition, comprising a development set of only 20 samples, our methodology strictly utilizes prompt engineering over traditional fine-tuning. Across the pipeline, each of the four Subtasks incorporates an LLM component. To quantify the performance gap between state-of-the-art proprietary models and locally deployable open-source alternatives, we systematically evaluated varying configurations of prompts and model architectures. Broadly, our prompts fall into three general categories: constraint-based instructions that enforce output structure and terminology, extraction-focused steps that isolate key information prior to generation, and rephrasing strategies that guide the model toward the desired output style. Depending on the complexity of the Subtask, we applied zero-shot prompting, few-shot ICL, lexical constraints, Task Decomposition, and CoT pipelines. We frequently merged these techniques to ensure robust adherence to task guidelines and mitigate hallucination. All models were executed with a temperature of 0.0 and a top-p of 0.95 to maximize deterministic behavior, though we acknowledge this may not be optimal for all model architectures evaluated.

\subsection{Subtask 1}

Our objective for Subtask 1 was to transform verbose patient-authored questions into concise clinical questions strictly under a 15-word limit. Each prompt received the patient question as input and instructed the model to generate a clinician interpreted query representing the medical information needed. The generated question was required to follow a strict JSON format containing a single field \textit{(``query'')}. 
To find the optimal balance between natural language generation and clinical accuracy, we systematically evaluated ten distinct prompt templates divided into two methodological strategies:

\begin{itemize}
    \item 
    \textbf{Direct Generation with Constraints (Prompts 1-7):} 
    This utilized a direct generation approach where the model was instructed to read the narrative and immediately synthesize a question. Prompt 1 established baseline zero-shot constraints (15-word limit, third-person perspective). Prompt 2 introduced lexical constraints to preserve exact medical terminology, to avoid paraphrasing, and Prompt 3 tested a few-shot alternative, providing 2 examples. Prompts 4 and 5 enforced verbatim extraction, explicitly forbidding the correction of misspellings to prevent model over-correction. Prompts 6 and 7 added structural rules enforcing the inclusion of medications/anatomical terms and a single continuous sentence format, respectively. However, stacking these negative and structural constraints within a single instruction frequently induced additional interference, causing the model to sporadically ignore rules or omit key terms.

    \item  
    \textbf{Task Decomposition (Prompts 8-10):} To mitigate constraint overload, we transitioned to an extract-then-generate pipeline. Prompt 8 forced the model to execute an intermediate step: explicitly identifying consecutive two- and three-word medical phrases. Prompt 9 refined this to isolate the primary clinical concern prior to formulating the query. Prompt 10 combined this sequential extraction with few-shot ICL to maximize terminology retention while adhering to brevity limits.

\end{itemize}

\subsection{Subtask 2}

Subtask 2 evaluates a system’s ability to identify clinical note sentences that provide evidence for answering a patient’s question. Due to limited development data, we framed this as a prompt-based inference task, instructing models to return the relevant sentence identifiers in a structured JSON format. 
Each prompt received the clinician question along with the numbered sentences of the clinical note excerpt, and the model was asked to select the identifiers of sentences that provide supporting evidence. We evaluated the following prompt formulations and ensemble strategy:

\begin{itemize}
    \item 
    \textbf{Sentence-Level Classification (Prompts 1-2):} Models evaluated each sentence independently, assigning labels (essential, supplementary, not relevant). This approach introduced instability across sentences and required complex post-hoc aggregation.

    \item 
    \textbf{Case-Level Evidence Selection (Prompts 3-8):} Prompts 3-6 received the entire excerpt and were instructed to output identifiers of relevant sentences directly, emphasizing a ``minimal and sufficient'' set. However, strict minimality often reduced recall by omitting partial supporting evidence. Prompts 7 and 8 introduced additional reasoning constraints (CoT) to improve transparency, but the increased prompt complexity destabilized output formatting.

    \item 
    \textbf{Recall- and Precision-Oriented Selection (Prompts 9-10):} To explore the trade-off between coverage and specificity, we tested two simplified formulations. Prompt 9 instructed the model to prefer precision over recall, selecting only the most directly relevant sentence identifiers, while Prompt 10 favoured recall over precision, selecting all sentence identifiers that could help answer the clinician's question to reduce the risk of missing relevant evidence. 

    \item 
    \textbf{Ensemble Aggregation:} To finalize predictions, we employed a majority voting ensemble across multiple LLMs \citep{dietterich2000ensemble}. Each model independently generated predictions, and sentence identifiers selected by a minimum of 2/3 models were included in the final output, reducing spurious selections while preserving consistently identified evidence.

\end{itemize}

\subsection{Subtask 3}
For Subtask 3, the goal was to synthesize a patient-friendly, evidence-grounded answer (maximum 75 words). Each prompt received the patient question, clinician question, and numbered clinical note sentences as input. 
Because this Subtask heavily evaluates factual consistency and adherence to the source text, our prompt engineering focused on preventing hallucination \cite{huang2025survey} through the following strategies:

\begin{itemize}

\item 
\textbf{Few-Shot with Explicit Constraints (Prompts 1-5, 8, 10):} We utilized a 1-shot learning approach, providing an annotated example from the task guidelines. We systematically varied the strictness of length instructions (e.g., ``exactly 3 to 5 sentences'' vs. ``maximum 75 words'') and paraphrasing allowances.

\item 
\textbf{Multi-Shot Demonstration (Prompt 6):} Expanded ICL by providing two distinct clinical examples, testing whether exposure to diverse specialties improved the model's ability to synthesize answers without exceeding the strict word count.

\item 
\textbf{Zero-Shot Abstractive Summarization (Prompt 7):} Removed all in-context examples to test baseline summarization capabilities, relying purely on instructional constraints to generate faithful, evidence-grounded answers.

\item 
\textbf{Implicit Chain-of-Thought / Pre-computation (Prompt 9):} Introduced a cognitive constraint requiring the LLM to silently identify relevant numbered sentences prior to generation. This anchored the model to an extractive mindset without polluting the final output with reasoning traces.

\item 
\textbf{High-Density Persona \& Negative Constraints (Prompt 11):} Shifted the system persona to a strict ``clinical auditor'' and densely stacked negative constraints (e.g., "Do NOT infer, generalize, explain, justify, speculate") to strongly discourage creative text generation.

\item 
\textbf{LLM-as-a-Judge:} To better generalize results across model-specific biases, we employed an LLM-as-a-Judge \cite{zheng2023judging} to select the single best candidate answer per case from a pool of outputs with the strongest model-prompt configurations. The judge evaluated candidates according to the following criteria, in priority order: (1) faithfulness: every claim must be supported by the clinical note sentences. (2) medical completeness: covering the key clinical information that answers the question. (3) terminology retention: preserving exact medical terms, procedure names, and diagnoses from the note. (4) conciseness: no filler or hedging, within the 75-word limit.

\end{itemize}

\subsection{Subtask 4}

Subtask 4 requires systems to link each sentence of a generated answer back to supporting sentences in the clinical note excerpt. 
Each prompt received the patient question, clinician question, the numbered clinical note sentences, and the generated answer sentences. To ensure accurate sentence-level labels, we preprocessed the texts by explicitly labeling them with structural identifiers prior to prompting. Specifically, each sentence in the source clinical note was prepended with a unique tag (e.g., [1], [2]), and each sentence in the generated answer was similarly tagged. In a variant, answer sentences were additionally prefixed with e.g. [N1], [N2] and [S1], [S2] tags to further disambiguate source and answer identifiers. Some configurations also employed a two-step verification pass, prompting the model to revisit any unaligned answer sentences after the initial output. We framed the model as a clinical evidence alignment expert tasked with generating strictly valid JSON under four core rules: only explicitly supported facts may be cited, references must contain the minimal set that contains all relevant information, unsupported sentences must return empty lists, and all answer sentences must be processed.


\begin{itemize}
    \item 
    \textbf{Zero-Shot Alignment (Prompts 1, 10):} Prompt 1 established the baseline with markdown-style rules and explicit-support constraints. Prompt 10 utilized a stricter formulation, explicitly forbidding inference and generalization. However, zero-shot models struggled to consistently differentiate direct evidence from loosely related context.
    
    \item 
    \textbf{Progressive Few-Shot Scaling (Prompts 2-9):} Prompt 2 introduced a 2-shot setting with clean alignment structures. Prompt 3 expanded to a 4-shot setting incorporating complex cases with high citation density and unsupported answers. Prompt 4 added a fifth conservative example to pull the model toward precise evidence linking. Prompt 5 tested a reordering of the Prompt 4 set to investigate and mitigate recency effects. Prompts 6 and 7 shifted back to 4-shot with alternative cases, while Prompts 8 and 9 extended this to 5-shot configurations, systematically varying example ordering and composition to identify the most effective sequence.

    \item 
    \textbf{Ensemble Aggregation:} Because all few-shot examples were drawn from the 20-case development set, individual prompt configurations risked overfitting. To improve generalization, we searched over candidate model-prompt configurations to construct an ensemble. 
\end{itemize}

\section{Results}

We employed a two-stage evaluation methodology across all four Subtasks. In the initial stage, we conducted extensive validation on the 20-case development set, evaluating a representative, though non-exhaustive, pool of state-of-the-art proprietary models (e.g., Gemini 3 Flash \cite{doshi2025gemini3flash}, Gemini 2.5 Flash \cite{comanici2025gemini}, Claude Sonnet 4.5 \cite{anthropic2025claudesonnet45}, GPT 4.1 \cite{openai2025gpt41}, Grok 4.1 Fast \cite{xai2025grok41fast}) alongside open-source alternatives (e.g., Llama 3.1 8B Instruct \cite{meta2024llama3}, Qwen 3 \cite{yang2025qwen3} and various domain finetunes \cite{Tran2024Bioinstruct}\footnote{\url{https://huggingface.co/khazarai/Bio-8B-it}}\footnote{\url{https://huggingface.co/Echelon-AI/Med-Qwen2-7B}}, MedGemma-27B \cite{sellergren2025medgemma}. We iteratively adjusted this baseline pool to incorporate newer, higher-performing models as they became available. This validation phase identified our top-performing configurations, which were subsequently applied to the hidden test set (comprising 47 samples for Subtasks 1-3 and 147 samples for Subtask 4) to generate the official results. To maximize competitive performance, our final submissions featured proprietary models rather than open-source alternatives; we acknowledge this reliance as a limitation of our work. Regarding computational resources, smaller open-source models (<70B parameters) were executed on a SLURM cluster utilizing up to two NVIDIA L4 (24GB) GPUs. All other models were accessed via API endpoints (OpenRouter), incurring a total cost of \$109.45 USD for the duration of the competition. The following subsections detail the validation findings and official performance for each Subtask, with a a complete set of validation experiments for Subtasks 1, 2, 3, and 4, including a brief evaluation of very large open-source models, is provided in the Appendix, for Subtask 1, 3, 4.

\subsection{Subtask 1}

\begin{figure}[htbp]
    \centering
    \includegraphics[width=1.0\linewidth]{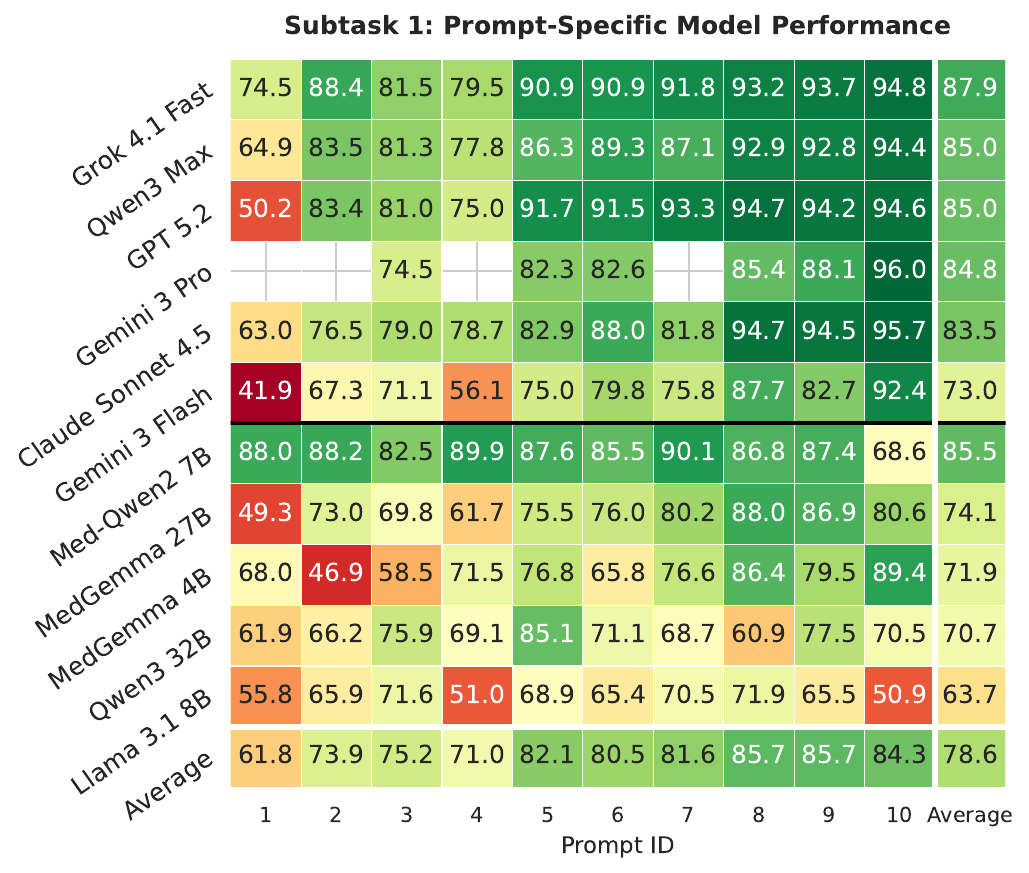}
    \caption{Subtask 1 Validation Results showing the different prompts over several different open and close source models. The scores represent the official evaluation metrics of the average of ROUGE, BERTScore, AlignScore, and MEDCON.}
    \label{fig:Subtask1_validation_results}
\end{figure}

Performance in Subtask 1 is evaluated using an average of four automatic text generation metrics: ROUGE \cite{lin-2004-rouge}, BERTScore \cite{Zhang2020BERTScore}, AlignScore \cite{zha-etal-2023-alignscore}, and MEDCON \cite{yim2023aci}.  Our validation results are presented in Figure~\ref{fig:Subtask1_validation_results}. Across the Direct Generation phase (Prompts 1-7), performance gradually improved as constraints were refined, though scores remained generally lower than those in the Task Decomposition phase (Prompts 8-10). Task Decomposition consistently achieved the highest average scores across model architectures, suggesting that breaking the task into sequential extraction steps is a highly effective strategy. Based on these validation results, Prompts 8, 9, and 10 were identified as the strongest configurations, with prompt 10 offering the  best performance for most proprietary models, indicating the advantage of using few shot examples. In general we note that most proprietary models offer alot more performance and robustness than most open source alternatives.

The official leaderboard results for Subtask 1 are presented in Table~\ref{tab:Subtask1_results}. Our primary submission utilized Claude Sonnet 4.5 with Prompt 10, combining the sequential extraction pipeline with 2-shot in-context learning (ICL) to maximize terminology retention within the 15-word limit. This approach achieved a score of 19.0, ranking 13th on the leaderboard. A secondary submission using GPT-5.2 with Prompt 8 (extract-then-generate) achieved a score of 16.6, indicating that the in-context examples in Prompt 10 provided a meaningful advantage. Notably, the substantial discrepancy between our internal validation scores and the official test results suggests a potential distribution shift in the test data, an overfitting effect to the highly constrained 20-case development set, or some internal errors on our part.

\begin{table}[htbp]
    \centering
    \begin{tabularx}{\columnwidth}{@{}l c C@{}}
        \toprule
        \textbf{Model / Team} & \textbf{Rank} & \textbf{Score} \\
        \midrule
        
        \textit{Our Submissions} \\
        Sonnet-4.5 Prompt 10 & 13 & 19.0 \\
        GPT-5.2 Prompt 8 & -- & 16.6 \\
        \midrule
        
        \textit{Leaderboard} \\
        Best Competitor & \textbf{1} & \textbf{31.2} \\
        Median       & 7 & 25.6 \\       
        \bottomrule
    \end{tabularx}
    \caption{Subtask 1 Official Results. The score represents the average of ROUGE, BERTScore, AlignScore, and MEDCON.}
        \label{tab:Subtask1_results}

\end{table}


\subsection{Subtask 2}

\begin{figure}[htbp]
    \centering
    \includegraphics[width=1.0\linewidth]{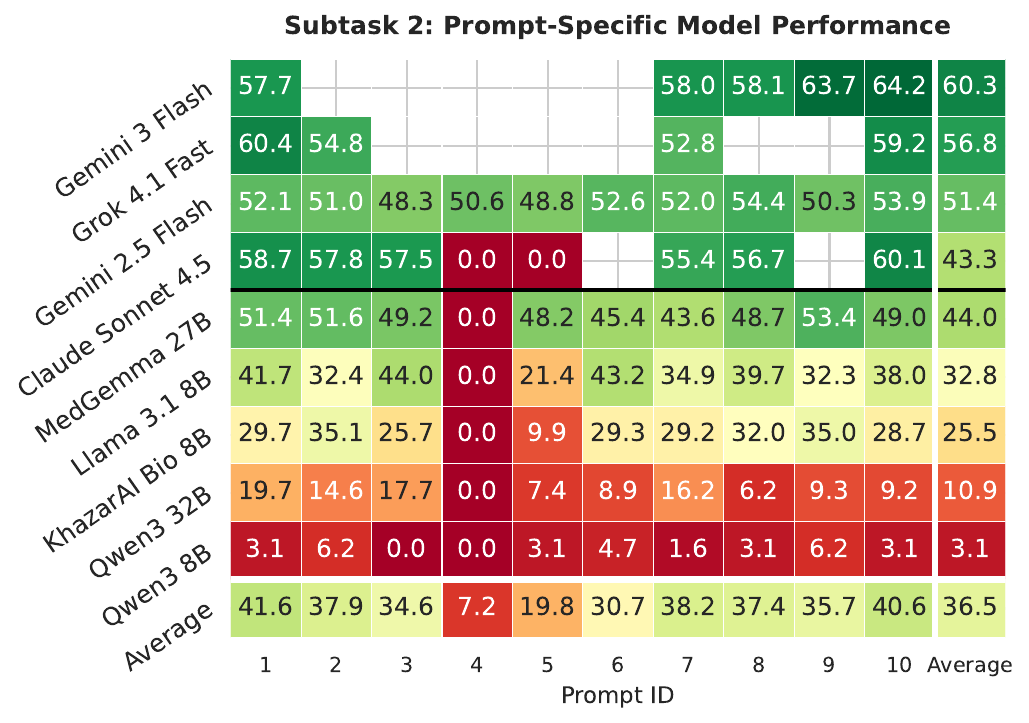}
    \caption{Subtask 2 Validation Results over the Strict Micro F1 metric}
    \label{fig:Subtask2_validation_results}
\end{figure}

Performance in Subtask 2 is evaluated using the Strict Micro F1 score to measure a system's ability to accurately identify the minimal and sufficient set of relevant evidence sentences. Validation results (Figure~\ref{fig:Subtask2_validation_results}) highlight a distinct performance gap between proprietary and open-source models. Proprietary models consistently achieved the highest scores, typically exceeding 50.0 Micro F1 and demonstrating relative stability across various prompt configurations. In contrast, open-source models produced substantially lower scores and exhibited higher sensitivity to prompt variations. Further we note many cases of complete failure (F1 score of zero), where the model/prompt configuration would generate responses which were unparseable. In terms of our prompting approaches, we noticed no clear winners in terms of prompts, with various models having various optimal prompts, with the exception of prompt 10, obtaining top performance for most propriety models.

Consequently, our final system relied on an ensemble of the strongest proprietary models. The primary submission, a majority voting ensemble comprising Gemini 3 Flash Preview, Grok 4.1 Fast, and Claude Sonnet 4.5 (all using prompt 10), achieved a Strict Micro F1 score of 58.8, ranking 11th on the leaderboard (Table~\ref{tab:Subtask2_results}). This performance is near the median system score (59.8) and remains competitive with the top-performing system (63.7). An alternative single-model submission (Grok 4 Fast Prompt 10) achieved a score of 56.0. The minimal variance between the ensemble and single-model runs demonstrates the stability of the prompting strategy, suggesting that a single model is highly reliable and rendering more computationally expensive ensembling strategies unnecessary.

\begin{table}[htbp]
\centering
\begin{tabularx}{\columnwidth}{@{}l c C@{}}
\toprule
\textbf{Model / Team} & \textbf{Rank} & \textbf{Strict Micro F1} \\
\midrule

\textit{Our Submissions} \\
Ensemble Top 3       & 11 & 58.8 \\
Grok 4 Fast (Prompt 10)      & -- & 56.0 \\

\midrule
\textit{Leaderboard} \\
Best Competitor  & \textbf{1} & \textbf{63.7} \\
Median       & 7/8 & 59.8 \\

\bottomrule
\end{tabularx}
\caption{Subtask 2 Performance: Strict Micro F1 on ArchEHR-QA.}
\label{tab:Subtask2_results}
\end{table}

\subsection{Subtask 3}

\begin{figure}[htbp]
    \centering
    \includegraphics[width=1.0\linewidth]{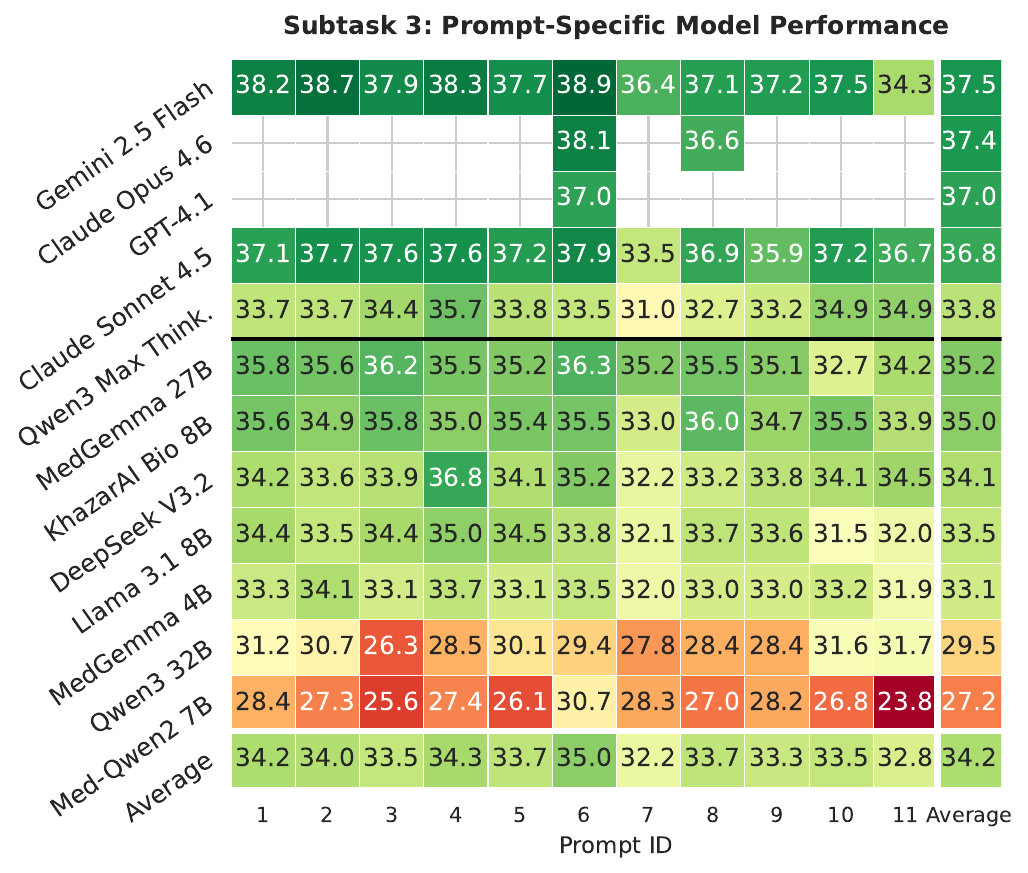}
    \caption{Subtask 3 Validation Results using an average score of the metrics: BLEU, ROUGE, SARI, BERTScore, AlignScore, and MEDCON}
    \label{fig:Subtask3_validation_results}
\end{figure}

Performance in Subtask 3 is evaluated using an average score over the metrics: BLEU, ROUGE, SARI \cite{xu-etal-2016-optimizing}, BERTScore, AlignScore, and MEDCON. Validation results (Figure \ref{fig:Subtask3_validation_results}) revealed that performance narrowly varied across all eleven prompts. This indicates that for patient-friendly answer generation, models are relatively robust to prompt variation, making model selection the primary driver of performance. Gemini 2.5 Flash and Claude Sonnet 4.5 consistently outperformed other models, while open-source models like MedGemma-27B and a domain fine-tuned Qwen 3-8B\footnote{\url{https://huggingface.co/khazarai/Bio-8B-it}} performed competitively, narrowing the performance gap observed in other Subtasks.

Given the marginal prompt-based variance, our submission strategy prioritized model diversity and high-quality demonstrations. Our first submission utilized an LLM-as-a-Judge (GPT-4.1) over the strongest model-prompt configurations:  Gemini 2.5 Flash, Claude Sonnet 4.5, and Claude Opus 4.6 on prompt 6. This approach ranked 3rd on the official leaderboard with a score of 35.6 (Table~\ref{tab:Subtask3_results}), closely trailing the top competitor (36.3). Single-model submissions also demonstrated strong performance: Gemini 2.5 Flash (Prompt 6, multi-shot) achieved 35.5, while Claude Sonnet 4.5 (Prompt 2, 1-shot explicit constraints) scored 33.9. This suggests that for this generation task, the LLM-as-a-Judge does not add much value relative to the overhead. The narrow 0.7-point gap between our best ensemble and the winning submission underscores the viability of prompt-based approaches for simplified medical text generation.

\begin{table}[htbp]
\centering
\begin{tabularx}{\columnwidth}{@{}l c C@{}}
\toprule
\textbf{Model / Team} & \textbf{Rank} & \textbf{Score} \\
\midrule
\textit{Our Submissions} \\
LLM-as-a-Judge 4 Models          & 3 & 35.6 \\
Gemini-2.5-Flash (Prompt 6)  & -- & 35.5 \\
Sonnet-4.5 (Prompt 2)       & -- & 33.9 \\
\midrule
\textit{Leaderboard} \\
Best Competitor & \textbf{1} & \textbf{36.3} \\
Leaderboard Median          & 7 & 32.9 \\

\bottomrule
\end{tabularx}
\caption{Subtask 3 Performance Comparison. The Overall metric is a composite score evaluating simplified medical text generation, incorporating BLEU, ROUGE-Lsum, SARI, BERTScore, AlignScore, and MEDCON (UMLS).}
\label{tab:Subtask3_results}
\end{table}

\subsection{Subtask 4}

\begin{figure}[htbp]
    \centering
    \includegraphics[width=1.0\linewidth]{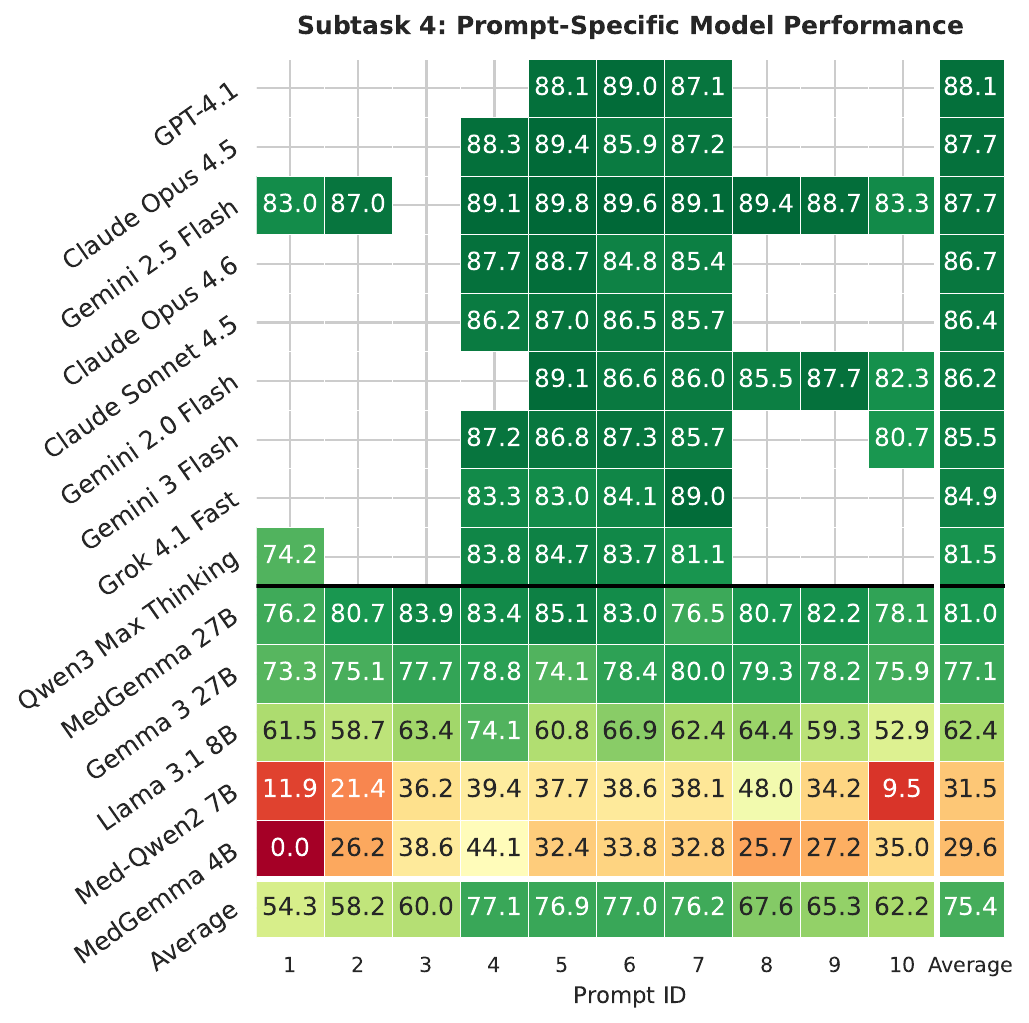}
    \caption{Subtask 4 Validation Results}
    \label{fig:Subtask4_validation_results}
\end{figure}

Performance in Subtask 4 is evaluated using sentence-level precision, recall, and Micro F1 metrics over predicted alignment links. Validation results (Figure \ref{fig:Subtask4_validation_results}) indicated that using few shot prompting (Prompts 2-9), often gave the open source models a clear benefit, with most models obtaining their best scores between prompts 5-7. However, because all few-shot examples were drawn from the evaluation development set, we believed that this might reflect partial overfitting. In general, proprietary models significantly outperformed open-source alternatives. Google's MedGemma-27B was a notable exception (average 81.0), though it still trailed the top proprietary models.

To maximize robustness, we utilized an ensemble search over the strongest model-prompt combinations (Prompts 5-7), with the final submissions using Prompts \textit{5} and \textit{6}. The final ensemble composition was selected based solely on Micro F1 performance on the 20-case development set, with no subsequent adjustment made prior to or following test set submission. Our primary submissions, a 3-model majority voting ensemble (Gemini 2.5 Flash, Claude Opus 4.6, Gemini 2.0 Flash) with \textit{Prompt 5}, achieved 1st place on the leaderboard with a Micro F1 of \textit{81.5} (Table \ref{tab:Subtask4_results}).  Our second submission used a 4-model ensemble combining Gemini 2.5 Flash and GPT-4.1, each evaluated with \textit{Prompts 5 and 6}, resulting in four model-prompt configurations and achieving \textit{81.3}. Finally, a strictly constrained zero-shot single-model configuration (Gemini 2.5 Flash, Prompt 10) achieved \textit{81.2}, demonstrating that carefully engineered zero-shot prompting can perform competitively with few-shot ensembles. Although the performance differences between our approaches is minimal, the margin between the 1st and 4th place submission was 1.2 Micro F1, indicating that these marginal changes affect the ranking of the systems. Furthermore, despite earlier concerns about potential overfitting due to few-shot examples drawn from the development set, the similarity between validation and official results suggests the overfitting was minimal.

\begin{table}[ht]
\centering
\begin{tabularx}{\columnwidth}{@{}l c C@{}}
\toprule
\textbf{Model / Team} & \textbf{Rank} & \textbf{Micro F1} \\
\midrule
\textit{Our Submissions} \\
Ensemble 3 models     & \textbf{1} & \textbf{81.5} \\
Ensemble 4 models  & -- & 81.3 \\
Gemini 2.5 Flash (Prompt 10)   & -- & 81.2 \\
\midrule
\textit{Leaderboard} \\
Best Competitor & {2} & {81.3} \\
Median       & 8/9 & 77.5 \\

\bottomrule
\end{tabularx}
\caption{Subtask 4 Performance: Micro F1 on ArchEHR-QA.}
\label{tab:Subtask4_results}
\end{table}

\section{Error Analysis}

\textbf{Subtask 1: Question Interpretation Error Analysis.}
The model occasionally failed to transform patient narratives into professional clinical queries. In Case 129, it generated an informal, query with lots of spelling errors (\textit{``Why is he not eatin feeling weak and shakey loosin weight...''}) rather than the required concise clinical formulation (\textit{“What explains his weakness, weight loss, and poor appetite...”}). Upon reflection this is a model directly following our instructions where we ask specifically to not rephrase or correct spelling errors.

\textbf{Subtask 2: Evidence Retrieval Error Analysis.}
Retrieval primarily suffered from over-selection and missed evidence. In over-selection cases (e.g., Case 19), the system successfully retrieved the gold annotations but included numerous false positives containing irrelevant procedural background due to an over-reliance on broad semantic similarity. Notably, our attempts to mitigate this using stricter filtering and ensemble methods have not fully resolved the issue. Conversely, in missed evidence cases (e.g., Case 20), the model hyper-focused on explicit diagnostic terminology, completely overlooking relevant sentences describing treatments. Addressing this requires non-trivial improvements in semantic reasoning.

\textbf{Subtask 3: Answer Generation Error Analysis.}
Generation errors typically involved medical hallucinations or context misinterpretation. In Case 21, the model correctly identified a patient's cirrhosis but hallucinated plausible yet entirely unsupported complications (\textit{portal gastropathy, Budd-Chiari syndrome}). In Case 29, it ignored the primary clinical note regarding deep vein thrombosis treatment, instead generating an unrelated discussion about Coumadin and alcohol use. These issues emphasize the need for stronger grounding mechanisms to tightly couple generation with the retrieved evidence.

\textbf{Subtask 4: Evidence Alignment Error Analysis.}
The model demonstrated difficulty with multi-evidence reasoning, where multiple context sentences jointly support a single claim. In Case 3, the model correctly cited one supporting sentence but missed a second complementary citation detailing warning symptoms, largely because the missing sentence utilized different terminology. Mitigating these incomplete citations will require prompting and retrieval strategies that explicitly encourage aggregating complementary evidence.

\section{Discussion}

In this work, we aimed to evaluate the limits of LLMs in the extreme low-resource biomedical setting of the ArchEHR competition, relying primarily on prompt engineering and model selection rather than supervised fine-tuning. Our findings highlight several key dynamics regarding model scaling, open-source viability, and the practical trade-offs of ensemble methodologies in clinical applications.

Addressing the viability of open-weights models for secure clinical deployment (RQ1), internal validation highlighted the impressive capabilities of open-source, domain-adapted models. Notably, MedGemma 3 27B performed exceptionally well, closely trailing large proprietary counterparts despite its relatively small parameter count. This shrinking performance gap is critical for the healthcare domain, where data privacy regulations often strictly prohibit sending sensitive patient data to external API endpoints. Locally deployable domain tuned open-source models are often the only option in these settings; however, our findings confirm that more work is needed to extract optimal performance from these smaller models.

Regarding the effectiveness of prompt engineering (RQ2), our validation phase revealed that the impact of specific techniques is inversely proportional to baseline model capability. State-of-the-art proprietary models generally demonstrated high robustness, yielding less performance variance across different prompt templates while maintaining higher relative performance, with several prompts often achieving similarly top-tier results. Conversely, open-source LLMs exhibited significant, model-dependent performance swings (e.g., the high variance between Qwen 3-32B and MedGemma architectures in Subtask 1, though Subtask 3 proved an exception). However, even for the most robust models, careful prompt tuning remained a necessity for enforcing strict structural constraints, such as rectifying the consistent JSON formatting failures observed in Subtask 2. Furthermore, it is important to note that poorly constructed prompts will still degrade the performance of proprietary models, as evidenced by Prompt 1 in Subtask 1. In general, regarding proprietary models, once a baseline of high performance is reached, further prompt engineering yields diminishing returns. This is aligned with the existing literature \cite{zhuo-etal-2024-prosa}.

 To maximize our competitive standing, we utilized ensembling techniques across the pipeline, including majority voting and LLM-as-a-judge frameworks. While these strategies consistently improved robustness and yielded positive results, the absolute performance gains were often marginal. In a shared task competition, fractional improvements are vital for leaderboard positioning. However, from a real-world clinical deployment perspective, these techniques introduce severe operational bottlenecks. Running multiple inferences per query effectively multiplies operational costs or latency by a factor of the ensemble size. For live, scalable health systems, the marginal gains of ensembling are not worth the resulting computational inefficiency.

Broadly, the ArchEHR competition demonstrates that LLMs are highly competitive in this low-resource clinical setting, though their efficacy varies significantly by task type. For Subtask 3 (answer generation), LLMs remain the natural and dominant choice, excelling at abstractive synthesis. Surprisingly, our prompt-based ensemble also achieved state-of-the-art results in Subtask 4 (evidence citation), which we initially hypothesized would be dominated by fine-tuned systems. Conversely, the leaderboard standings for Subtask 2 suggest that other approaches may be more optimal over prompt-based LLMs for pure evidence-span retrieval, if the best competitor uses other techniques. Finally, we note that our official Subtask 1 performance was an anomaly compared to our internal validation; we believe there remains substantial, untapped potential for LLM-driven query transformation in this space.

\section{Conclusion}

This study investigated the efficacy of LLMs in the extreme low-resource clinical setting of the ArchEHR competition. By systematically evaluating zero-shot, few-shot, and CoT prompting strategies alongside ensemble methodologies, we demonstrated that state-of-the-art LLMs can achieve highly competitive performance across complex biomedical NLP tasks without any supervised fine-tuning. Notably, our prompt-based ensemble approach achieved first place in Subtask 4 (evidence citation alignment) and third place in Subtask 3 (patient-friendly answer generation), proving that LLMs are naturally suited for both abstractive synthesis and strict evidence-linking. Conversely, our lower performance in Subtasks 1 and 2 indicates that pure evidence-span retrieval and strict query formulation remain challenging for generative, prompt-only architectures.

Our findings also highlight critical dynamics for the real-world deployment of clinical AI. While massive proprietary models exhibited strong robustness to prompt variations, open-source models, particularly MedGemma 3 27B, demonstrated highly competitive capabilities with certain prompts. This narrowing performance gap is a crucial development for healthcare environments bound by strict data privacy frameworks like GDPR, where local deployment is mandatory. Finally, while ensembling techniques such as majority voting and LLM-as-a-judge provided the marginal gains necessary for competitive leaderboard positioning, their associated computational costs and latency make them inefficient for scalable, real-world clinical deployment. 


\section{Limitations}

While our work showed good results, this study has several notable limitations. First, due to the extreme low-resource nature of the shared task (only 20 development cases), there is a persistent risk that our few-shot prompts and ensemble configurations are partially overfit to the development distribution, despite our efforts to utilize varying case structures. This impacts the conclusions drawn from the validation results. Additionally, our uniform use of temperature 0.0 and top-p of 0.95 across all models, while intended to maximize deterministic behavior, may not be optimal for all architectures evaluated, particularly reasoning-oriented models that are designed to benefit from some degree of stochasticity. 

Second, our evaluation of proprietary LLMs was not exhaustive, and was not run on all prompt configurations. The financial constraints associated with commercial API costs limited our ability to perform evaluate every prompt variation across all available proprietary models. Consequently, the optimal prompt-model combinations reported may reflect a local maximum rather than absolute peak performance.

Furthermore, our system heavily relies on proprietary, closed-weights models (e.g., Gemini and Claude) accessed via API endpoints to achieve top-tier performance. This introduces reproducibility challenges, as the underlying model weights may be silently updated by the providers. And in GDPR strict scenarios this cannot be permitted. Finally, the evaluation was restricted to English clinical notes; the efficacy of these prompting strategies on multilingual or non-English EHRs remains unverified.

\section{Ethical Considerations}
The deployment of generative LLMs in clinical settings carries profound ethical implications, primarily concerning patient safety and data privacy. Generative models are inherently prone to hallucination. While our system achieved high precision in evidence alignment (Subtask 4) and grounding (Subtask 3), the risk of clinical hallucination cannot be entirely eliminated through prompt engineering alone. Therefore, systems like ours must strictly be deployed as ``human-in-the-loop'' assistive tools to reduce clinician cognitive load, rather than as autonomous diagnostic or patient-facing decision-makers.

Additionally, our highest-performing submissions rely on sending clinical text to external APIs. In real-world healthcare environments governed by privacy frameworks such as HIPAA and GDPR, transmitting sensitive Protected Health Information to external servers is often legally prohibited. While we demonstrated that local, open-weights models like MedGemma 3 27B show strong promise, bridging the performance gap between secure local models and proprietary systems remains a critical ethical and technical imperative for the field.

\section{Acknowledgments}
This work was funded by FEDER - Fundo Europeu de Desenvolvimento Regional funds through Programa Regional do Centro, within project CENTRO2030-FEDER-02595400 and by the Foundation for Science and Technology (FCT) through the contract \url{https://doi.org/10.54499/UID/00127/2025}. Richard A. A. Jonker is funded by the FCT doctoral grant PRT/BD/154792/2023, 
with DOI identifier \url{https://doi.org/10.54499/PRT/BD/154792/2023}. Alexander Christiansen and Alexandros Maniatis are funded by the Danish Data Science Academy (DDSA) under grant numbers 2026-6484 and 2026-6480, respectively.

\section{Bibliographical References}\label{sec:reference}

\bibliographystyle{lrec2026-natbib}
\bibliography{refs}

@inproceedings{wei2022cot,
  title={Chain-of-Thought Prompting Elicits Reasoning in Large Language Models},
  author={Wei, Jason and Wang, Xuezhi and Schuurmans, Dale and Bosma, Maarten and Ichter, Brian and Xia, Fei and Chi, Ed H. and Le, Quoc V. and Zhou, Denny},
  booktitle={Advances in Neural Information Processing Systems (NeurIPS)},
  year={2022}
}

@article{brown2020language,
  title={Language models are few-shot learners},
  author={Brown, Tom B and Mann, Benjamin and Ryder, Nick and Subbiah, Melanie and Kaplan, Jared and Dhariwal, Prafulla and Neelakantan, Arvind and Shyam, Pranav and Sastry, Girish and Askell, Amanda and others},
  journal={Advances in neural information processing systems},
  volume={33},
  pages={1877--1901},
  year={2020}
}

@inproceedings{meng2023tuning,
  title={Tuning language models as training data generators for augmentation-enhanced few-shot learning},
  author={Meng, Yu and Michalski, Martin and Huang, Jiaxin and Zhang, Yu and Abdelzaher, Tarek and Han, Jiawei},
  booktitle={International Conference on Machine Learning},
  pages={24457--24477},
  year={2023},
  organization={PMLR}
}

@article{sahoo2024systematic,
  title={A systematic survey of prompt engineering in large language models: Techniques and applications},
  author={Sahoo, Pranab and Singh, Ayush Kumar and Saha, Sriparna and Jain, Vinija and Mondal, Samrat and Chadha, Aman},
  journal={arXiv preprint arXiv:2402.07927},
  volume={1},
  year={2024}
}

@article{chen2025unleashing,
  title={Unleashing the potential of prompt engineering for large language models},
  author={Chen, Banghao and Zhang, Zhaofeng and Langren{\'e}, Nicolas and Zhu, Shengxin},
  journal={Patterns},
  volume={6},
  number={6},
  year={2025},
  publisher={Elsevier}
}

@inproceedings{schall-de-melo-2025-hidden,
    title = "The Hidden Cost of Structure: How Constrained Decoding Affects Language Model Performance",
    author = "Schall, Maximilian  and
      de Melo, Gerard",
    editor = "Angelova, Galia  and
      Kunilovskaya, Maria  and
      Escribe, Marie  and
      Mitkov, Ruslan",
    booktitle = "Proceedings of the 15th International Conference on Recent Advances in Natural Language Processing - Natural Language Processing in the Generative AI Era",
    month = sep,
    year = "2025",
    address = "Varna, Bulgaria",
    publisher = "INCOMA Ltd., Shoumen, Bulgaria",
    url = "https://aclanthology.org/2025.ranlp-1.124/",
    pages = "1074--1084"
}

@inproceedings{zhou2023leasttomost,
title={Least-to-Most Prompting Enables Complex Reasoning in Large Language Models},
author={Denny Zhou and Nathanael Sch{\"a}rli and Le Hou and Jason Wei and Nathan Scales and Xuezhi Wang and Dale Schuurmans and Claire Cui and Olivier Bousquet and Quoc V Le and Ed H. Chi},
booktitle={The Eleventh International Conference on Learning Representations },
year={2023},
url={https://openreview.net/forum?id=WZH7099tgfM}
}

@article{thirunavukarasu2023large,
  title={Large language models in medicine},
  author={Thirunavukarasu, Arun James and Ting, Darren Shu Jeng and Elangovan, Kabilan and Gutierrez, Laura and Tan, Ting Fang and Ting, Daniel Shu Wei},
  journal={Nature medicine},
  volume={29},
  number={8},
  pages={1930--1940},
  year={2023},
  publisher={Nature Publishing Group US New York}
}

@inproceedings{dietterich2000ensemble,
  title={Ensemble methods in machine learning},
  author={Dietterich, Thomas G},
  booktitle={International workshop on multiple classifier systems},
  pages={1--15},
  year={2000},
  organization={Springer}
}

@article{huang2025survey,
  title={A survey on hallucination in large language models: Principles, taxonomy, challenges, and open questions},
  author={Huang, Lei and Yu, Weijiang and Ma, Weitao and Zhong, Weihong and Feng, Zhangyin and Wang, Haotian and Chen, Qianglong and Peng, Weihua and Feng, Xiaocheng and Qin, Bing and others},
  journal={ACM Transactions on Information Systems},
  volume={43},
  number={2},
  pages={1--55},
  year={2025},
  publisher={ACM New York, NY}
}

@article{zheng2023judging,
  title={Judging llm-as-a-judge with mt-bench and chatbot arena},
  author={Zheng, Lianmin and Chiang, Wei-Lin and Sheng, Ying and Zhuang, Siyuan and Wu, Zhanghao and Zhuang, Yonghao and Lin, Zi and Li, Zhuohan and Li, Dacheng and Xing, Eric and others},
  journal={Advances in neural information processing systems},
  volume={36},
  pages={46595--46623},
  year={2023}
}

@misc{doshi2025gemini3flash,
  title={Gemini 3 Flash: frontier intelligence built for speed},
  author={Doshi, Tulsee and {The Gemini Team}},
  year={2025},
  month={Dec},
  howpublished={Google The Keyword Blog},
  url={https://blog.google/products-and-platforms/products/gemini/gemini-3-flash/},
  note={Accessed: 2026-03-10}
}

@article{comanici2025gemini,
  title={Gemini 2.5: Pushing the frontier with advanced reasoning, multimodality, long context, and next generation agentic capabilities},
  author={Comanici, Gheorghe and Bieber, Eric and Schaekermann, Mike and Pasupat, Ice and Sachdeva, Noveen and Dhillon, Inderjit and Blistein, Marcel and Ram, Ori and Zhang, Dan and Rosen, Evan and others},
  journal={arXiv preprint arXiv:2507.06261},
  year={2025}
}

@misc{anthropic2025claudesonnet45,
  title={Introducing Claude Sonnet 4.5},
  author={{Anthropic}},
  year={2025},
  month={Sep},
  howpublished={Anthropic News},
  url={https://www.anthropic.com/news/claude-sonnet-4-5},
  note={Accessed: 2026-03-10}
}

@misc{xai2025grok41fast,
  title={Grok 4.1 Fast and Agent Tools API},
  author={{xAI}},
  year={2025},
  month={Nov},
  howpublished={xAI News},
  url={https://x.ai/news/grok-4-1-fast},
  note={Accessed: 2026-03-10}
}

@misc{openai2025gpt41,
  title={Introducing GPT-4.1 in the API},
  author={{OpenAI}},
  year={2025},
  month={Apr},
  howpublished={OpenAI Blog},
  url={https://openai.com/index/gpt-4-1/},
  note={Accessed: 2026-03-10}
}

@article{meta2024llama3,
  title={The Llama 3 Herd of Models},
  author={{Llama Team, AI @ Meta}},
  journal={arXiv preprint arXiv:2407.21783},
  year={2024},
  url={https://arxiv.org/abs/2407.21783}
}

@article{yang2025qwen3,
  title={Qwen3 technical report},
  author={{Qwen Team}},
  journal={arXiv preprint arXiv:2505.09388},
  year={2025}
}

@article{Tran2024Bioinstruct,
    author = {Tran, Hieu and Yang, Zhichao and Yao, Zonghai and Yu, Hong},
    title = {BioInstruct: instruction tuning of large language models for biomedical natural language processing},
    journal = {Journal of the American Medical Informatics Association},
    year = {2024},
    doi = {10.1093/jamia/ocae122}
}

@article{sellergren2025medgemma,
  title={MedGemma Technical Report},
  author={Sellergren, Andrew and Kazemzadeh, Sahar and Jaroensri, Tiam and Kiraly, Atilla and Traverse, Madeleine and Kohlberger, Timo and Xu, Shawn and Jamil, Fayaz and Hughes, Cían and Lau, Charles and others},
  journal={arXiv preprint arXiv:2507.05201},
  year={2025}
}

@inproceedings{lin-2004-rouge,
    title = "{ROUGE}: A Package for Automatic Evaluation of Summaries",
    author = "Lin, Chin-Yew",
    booktitle = "Text Summarization Branches Out",
    month = jul,
    year = "2004",
    address = "Barcelona, Spain",
    publisher = "Association for Computational Linguistics",
    url = "https://aclanthology.org/W04-1013/",
    pages = "74--81"
}

@inproceedings{Zhang2020BERTScore,
title={BERTScore: Evaluating Text Generation with BERT},
author={Tianyi Zhang and Varsha Kishore and Felix Wu and Kilian Q. Weinberger and Yoav Artzi},
booktitle={International Conference on Learning Representations},
year={2020},
url={https://openreview.net/forum?id=SkeHuCVFDr}
}

@inproceedings{zha-etal-2023-alignscore,
    title = "{A}lign{S}core: Evaluating Factual Consistency with A Unified Alignment Function",
    author = "Zha, Yuheng  and
      Yang, Yichi  and
      Li, Ruichen  and
      Hu, Zhiting",
    editor = "Rogers, Anna  and
      Boyd-Graber, Jordan  and
      Okazaki, Naoaki",
    booktitle = "Proceedings of the 61st Annual Meeting of the Association for Computational Linguistics (Volume 1: Long Papers)",
    month = jul,
    year = "2023",
    address = "Toronto, Canada",
    publisher = "Association for Computational Linguistics",
    url = "https://aclanthology.org/2023.acl-long.634/",
    doi = "10.18653/v1/2023.acl-long.634",
    pages = "11328--11348"
}

@article{yim2023aci,
  title={Aci-bench: a novel ambient clinical intelligence dataset for benchmarking automatic visit note generation},
  author={Yim, Wen-wai and Fu, Yujuan and Ben Abacha, Asma and Snider, Neal and Lin, Thomas and Yetisgen, Meliha},
  journal={Scientific data},
  volume={10},
  number={1},
  pages={586},
  year={2023},
  publisher={Nature Publishing Group UK London}
}

@article{xu-etal-2016-optimizing,
    title = "Optimizing Statistical Machine Translation for Text Simplification",
    author = "Xu, Wei  and
      Napoles, Courtney  and
      Pavlick, Ellie  and
      Chen, Quanze  and
      Callison-Burch, Chris",
    editor = "Lee, Lillian  and
      Johnson, Mark  and
      Toutanova, Kristina",
    journal = "Transactions of the Association for Computational Linguistics",
    volume = "4",
    year = "2016",
    address = "Cambridge, MA",
    publisher = "MIT Press",
    url = "https://aclanthology.org/Q16-1029/",
    doi = "10.1162/tacl_a_00107",
    pages = "401--415"
}

@inproceedings{zhuo-etal-2024-prosa,
    title = "{P}ro{SA}: Assessing and Understanding the Prompt Sensitivity of {LLM}s",
    author = "Zhuo, Jingming  and
      Zhang, Songyang  and
      Fang, Xinyu  and
      Duan, Haodong  and
      Lin, Dahua  and
      Chen, Kai",
    editor = "Al-Onaizan, Yaser  and
      Bansal, Mohit  and
      Chen, Yun-Nung",
    booktitle = "Findings of the Association for Computational Linguistics: EMNLP 2024",
    month = nov,
    year = "2024",
    address = "Miami, Florida, USA",
    publisher = "Association for Computational Linguistics",
    url = "https://aclanthology.org/2024.findings-emnlp.108/",
    doi = "10.18653/v1/2024.findings-emnlp.108",
    pages = "1950--1976"
}

@inproceedings{soni-etal-2026-archehr-qa,
  title = "Overview of the ArchEHR-QA 2026 Shared Task on Grounded Question Answering from Electronic Health Records",
  author = "Soni, Sarvesh and Demner-Fushman, Dina",
  booktitle = "Proceedings of the Third Workshop on Patient-Oriented Language Processing (CL4Health)",
  year = "2026",
  address = "Palma, Mallorca (Spain)",
  publisher = "ELRA",
}

@article{soni-demner-fushman-2026-dataset,
  title = {A Dataset for Addressing Patient's Information Needs related to Clinical Course of Hospitalization},
  author = {Soni, Sarvesh and Demner-Fushman, Dina},
  journal = {Scientific Data},
  year = {2026},
  month = {02},
  date = {2026-02-25},
  doi = {10.1038/s41597-026-06639-z},
  url = {https://doi.org/10.1038/s41597-026-06639-z},
  issn = {2052-4463}
}

@article{clinicalbert,
author = {Kexin Huang and Jaan Altosaar and Rajesh Ranganath},
title = {ClinicalBERT: Modeling Clinical Notes and Predicting Hospital Readmission},
year = {2019},
journal = {arXiv:1904.05342},
}

@article{singhal2023large,
  title={Large language models encode clinical knowledge},
  author={Singhal, Karan and Azizi, Shekoofeh and Tu, Tao and Mahdavi, S Sara and Wei, Jason and Chung, Hyung Won and Scales, Nathan and Tanwani, Ajay and Cole-Lewis, Heather and Pfohl, Stephen and others},
  journal={Nature},
  volume={620},
  number={7972},
  pages={172--180},
  year={2023},
  publisher={Nature Publishing Group UK London}
}

@misc{gdpr2016,
  title = {Regulation ({EU}) 2016/679 of the {European} {Parliament} and of the {Council} of 27 {April} 2016 on the protection of natural persons with regard to the processing of personal data and on the free movement of such data, and repealing {Directive} 95/46/{EC} ({General} {Data} {Protection} {Regulation})},
  author = {{European Union}},
  howpublished = {Official Journal of the European Union, L 119, 1-88},
  year = {2016},
  url = {https://eur-lex.europa.eu/eli/reg/2016/679/oj}
}

@misc{hipaa1996,
  title = {Health {Insurance} {Portability} and {Accountability} {Act} of 1996 ({HIPAA})},
  author = {{104th United States Congress}},
  howpublished = {Public Law 104-191, 110 Stat. 1936},
  year = {1996},
  url = {https://www.govinfo.gov/app/details/PLAW-104publ191}
}


\appendix
\section{Additional Experiments}

\begin{figure*}[htbp]
    \centering
    \includegraphics[width=1.0\linewidth]{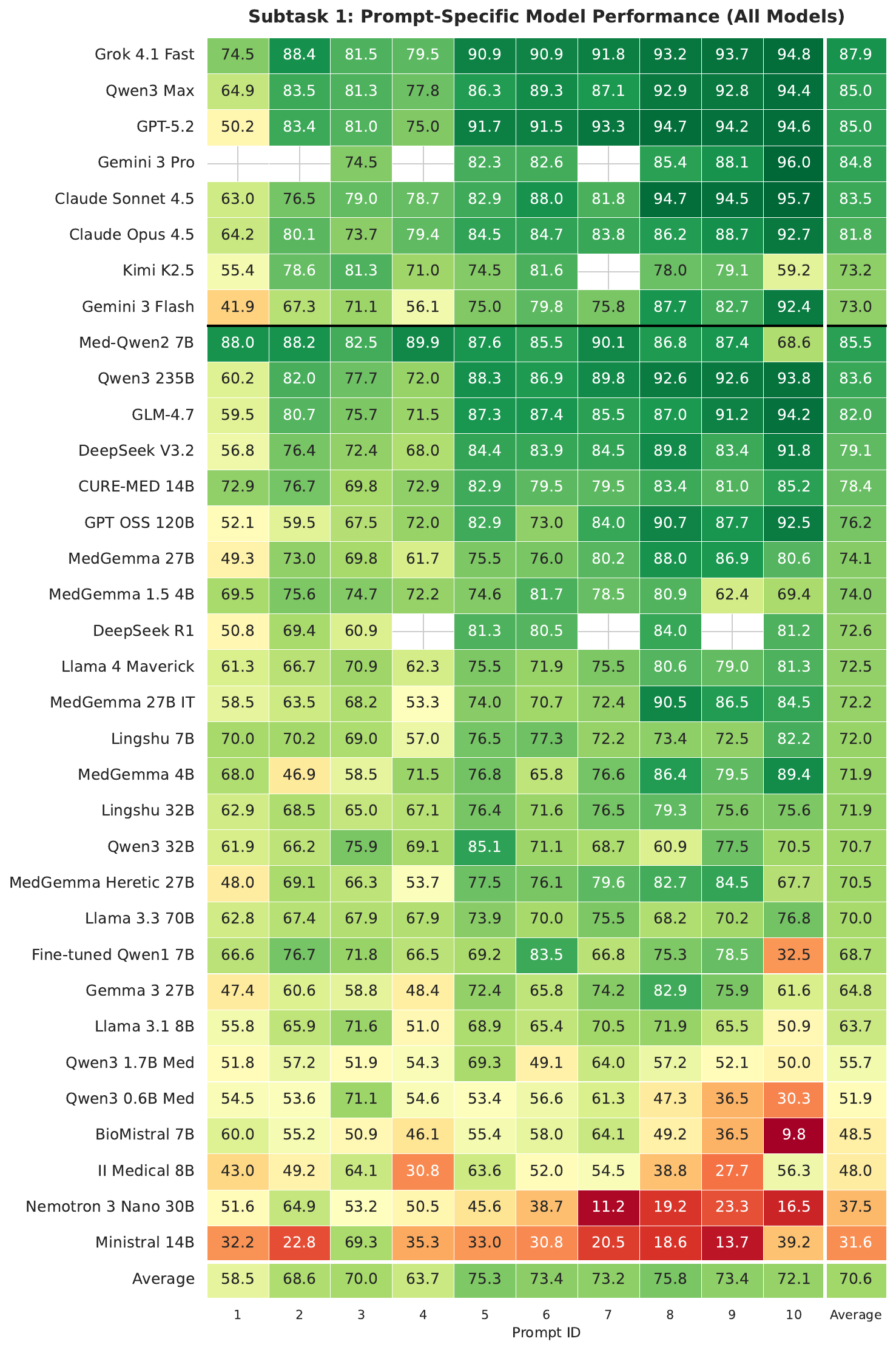}
    \caption{Subtask 1 Validation Results (All Models)}
    \label{fig:Subtask1_validation_results_appendix}
\end{figure*}
\begin{figure*}[htbp]
    \centering
    \includegraphics[width=1.0\linewidth]{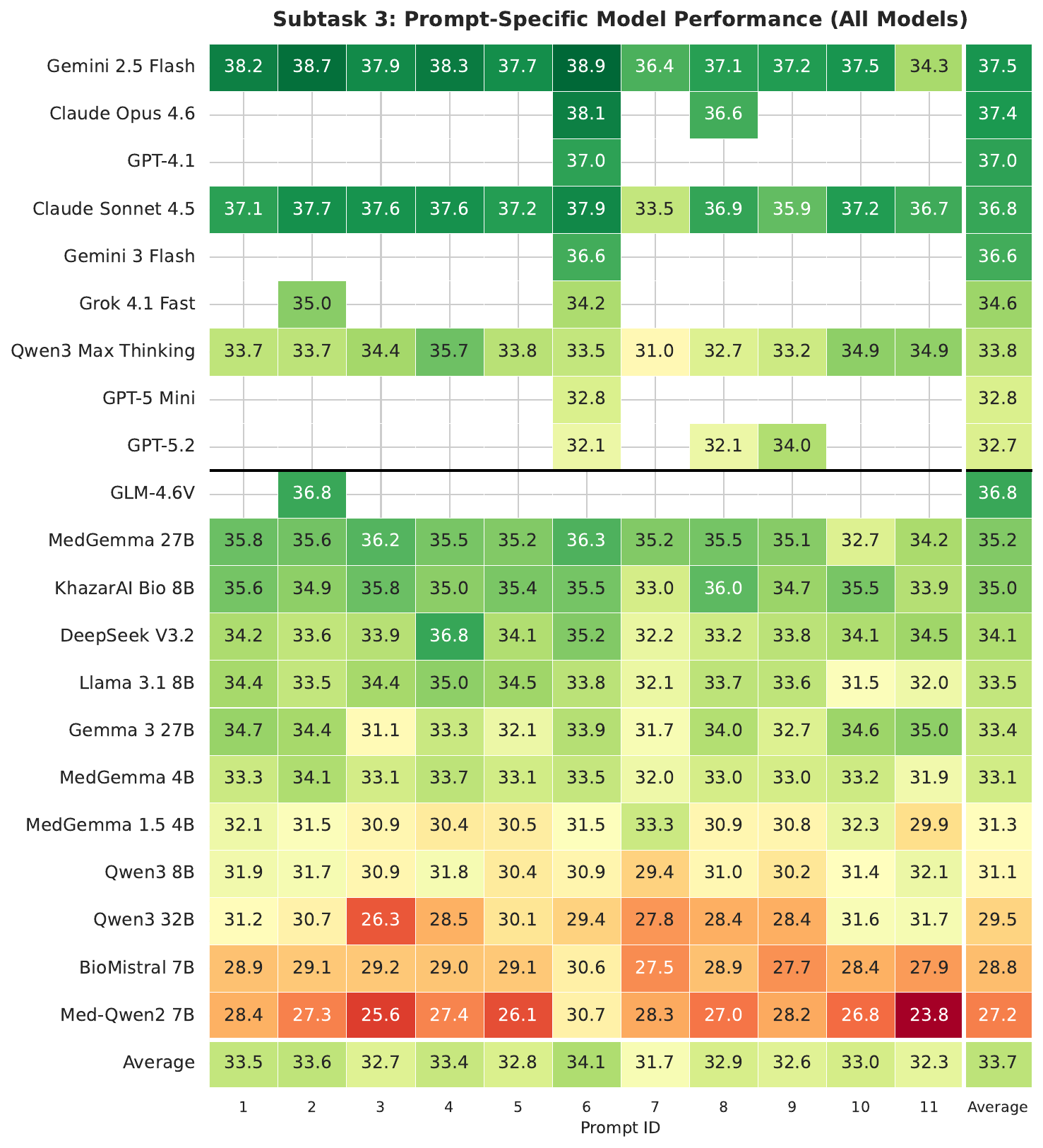}
    \caption{Subtask 3 Validation Results (All Models)}
    \label{fig:Subtask3_validation_results_appendix}
\end{figure*}
\begin{figure*}[htbp]
    \centering
    \includegraphics[width=1.0\linewidth]{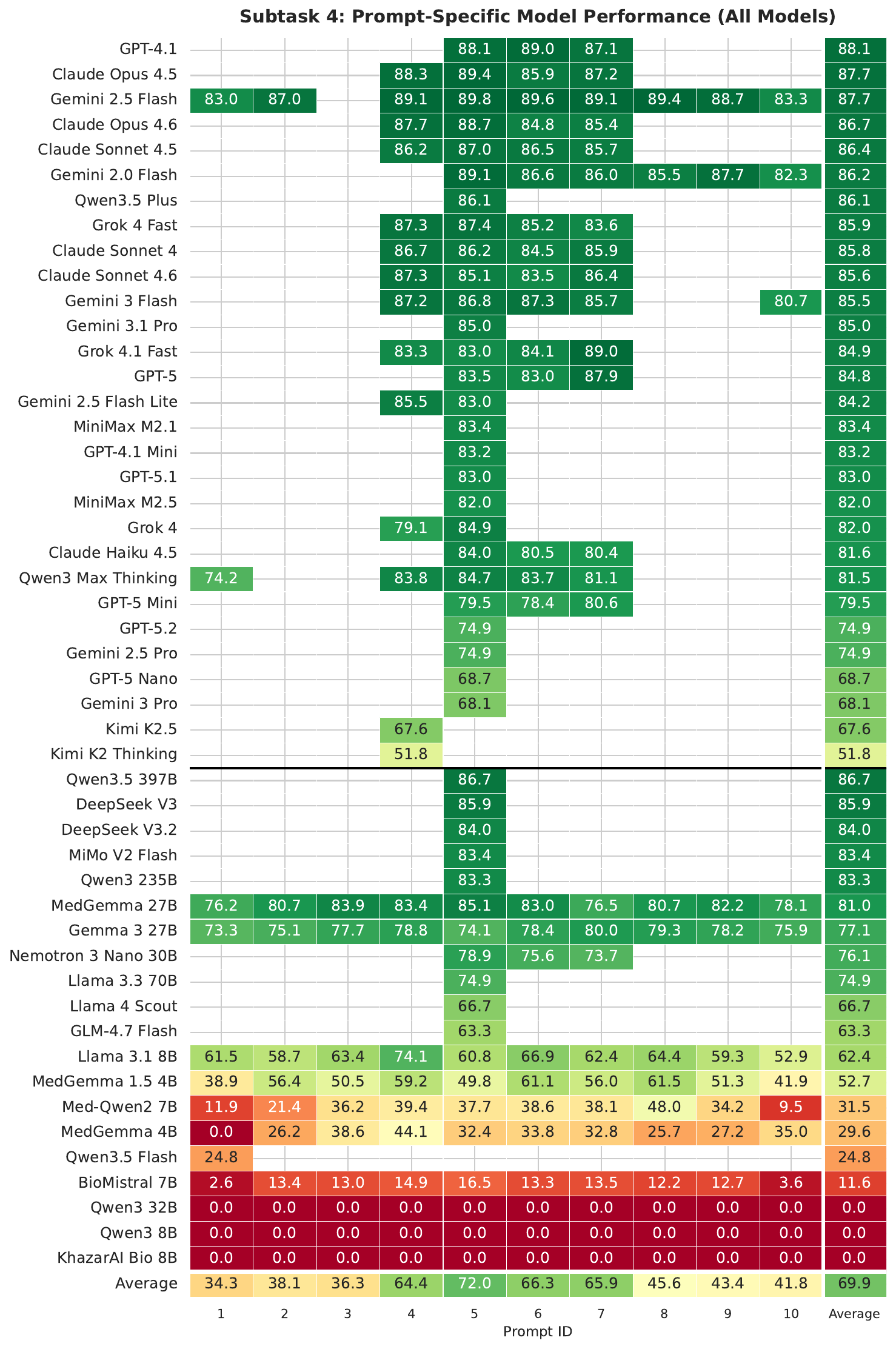}
    \caption{Subtask 4 Validation Results (All Models)}
    \label{fig:Subtask4_validation_results_appendix}
\end{figure*}

In this appendix, we present the full set of experiments conducted on the development set for Subtasks 1, 3, and 4 (Figures \ref{fig:Subtask1_validation_results_appendix}, \ref{fig:Subtask3_validation_results_appendix}, and \ref{fig:Subtask4_validation_results_appendix}, respectively). Notably, these extended validation results demonstrate that very large open-source models (such as Qwen 3 235B, GLM4.7, and Deepseek V3.2) performed exceptionally well, achieving results highly competitive with proprietary models. However, we chose not to emphasize these models in the main text due to their prohibitive local hardware requirements. Running these models locally requires substantial compute resources (often exceeding 140GB of VRAM), which is impractical for most standard biomedical deployments. Because of this limitation, we accessed these open-source models via API endpoints during testing; however, for API-based applications, we opted to focus our primary analysis on established proprietary models. For these practical reasons, we did not investigate these massive open-source models further in the main study.

\end{document}